\pdfoutput=1

\documentclass[10pt, a4paper]{article}

\usepackage{lrec-coling2024} 

\usepackage{times}
\usepackage{latexsym}
\usepackage{graphicx}

\usepackage[T1]{fontenc}

\usepackage[utf8]{inputenc}

\usepackage{microtype}

\usepackage{xspace}
\usepackage{booktabs}
\usepackage{graphicx}
\usepackage{enumitem}
\usepackage{xcolor}
\usepackage{tikz-dependency}
\usepackage{tikz}
\usetikzlibrary{positioning}
\usepackage{url}
\usepackage{amssymb}
\usepackage{float}
\usepackage{todonotes}
\usepackage{multirow}

\usepackage{fontawesome5}

\setlist[description]{leftmargin=\parindent,labelindent=0pt}

\newcommand{\enquote}[1]{``#1''}

\newcommand{\sref}[1]{Section~\ref{#1}}
\newcommand{\aref}[1]{Appendix~\ref{#1}}

\newcommand{\tref}[1]{Table~\ref{#1}}
\newcommand{\fref}[1]{Figure~\ref{#1}}

\usepackage{soul}
\sethlcolor{white} 

\newcommand{\corpusName}{AnnoCTR\xspace} 
\newcommand{\mitreAttack}{MITRE ATT\&CK\xspace}
\newcommand{\mitreAttackShort}{ATT\&CK\xspace}
\newcommand{\numAnnotatedReports}{120\xspace}

\newcommand{\cysecSpecificAnnotations}{13,244\xspace} 
\newcommand{\totalAnnotations}{20,855\xspace} 
\newcommand{\numSentencesCysecPart}{12,179\xspace} 

\newcommand{\sd}[1]{{\tiny $_{\pm #1}$}}

\setlist[description]{leftmargin=\parindent,labelindent=0pt}

\title{\corpusName: A Dataset for Detecting and Linking Entities,\\ Tactics, and Techniques in Cyber Threat Reports}

\name{Lukas Lange$^{\ast}$ \hspace{2mm} Marc Müller\textsuperscript{\#} \hspace{2mm} Ghazaleh Haratinezhad Torbati$^{\dagger}$ \\ {\bf \large Dragan Milchevski$^{\ast}$ \hspace{2mm}
 Patrick Grau$^{\ddagger}$ \hspace{2mm} Subhash Pujari$^{\ast}$ \hspace{2mm} Annemarie Friedrich$^{\mathsection}$}}

\address{
    $^{\ast}$Bosch Center for Artificial Intelligence, Renningen, Germany \\ 
    {\{lukas.lange, dragan.milchesvki, subhash.pujari\}@de.bosch.com} \\ [0.3em]
    $^{\dagger}$Max Planck Institute for Informatics, Saarbrücken, Germany \\ [0.3em]
    \textsuperscript{\#}Hochschule der Medien, Stuttgart, Germany \hspace{3mm} $^{\ddagger}$Robert Bosch GmbH, Stuttgart, Germany \\ [0.3em]  
    $^{\mathsection}$Universität Augsburg, Augsburg, Germany \\ 
    annemarie.friedrich@informatik.uni-augsburg.de
}

\abstract{
Monitoring the threat landscape to be aware of actual or potential attacks is of utmost importance to cybersecurity professionals.
Information about cyber threats is typically distributed using natural language reports.
Natural language processing can help with managing this large amount of unstructured information, yet to date, the topic has received little attention.
With this paper, we present AnnoCTR, a new CC-BY-SA-licensed dataset of cyber threat reports.
The reports have been annotated by a domain expert with named entities, temporal expressions, and cybersecurity-specific concepts including implicitly mentioned techniques and tactics.
Entities and concepts are linked to Wikipedia and the \mitreAttack knowledge base, the most widely-used taxonomy for classifying types of attacks. 
Prior datasets linking to \mitreAttack either provide a single label per document or annotate sentences out-of-context; our dataset annotates entire documents in a much finer-grained way.
In an experimental study, we model the annotations of our dataset using state-of-the-art neural models. 
In our few-shot scenario, we find that for identifying the \mitreAttack concepts that are mentioned explicitly or implicitly in a text, 
concept descriptions from \mitreAttack are an effective source for training data augmentation.
\\ \newline \Keywords{Cybersecurity, Concept Detection, Named Entity Recognition, Entity Linking}
}

\begin{document}
\maketitleabstract

\section{Introduction}
\label{sec:intro}

Cyber Threat Intelligence (CTI) necessitates collecting evidence-based knowledge about cyber threats to proactively defend against cyber attacks.
Cyber Threat Reports (CTRs), which are usually provided by professional CTI vendors, are unstructured text documents that describe threat-related information such as tactics, techniques, actors, tools, types of systems as well as geographic regions, political entities, or targeted industries.
Retrieving and analysing information from CTRs is a tedious and time-consuming yet usually time-critical task \citep{sarhan2021open,rahman2021what}.
Obtaining clean labeled data that ensures replication, validation and extensions of CTI studies constitutes a major technical challenge \citep{rahman2021what}.

\begin{figure}[]
    \centering
    \footnotesize

    \tikzset{
    every node/.style={inner sep=1pt,minimum height=4mm,rounded corners=.5mm,font=\fontsize{8}{8}\selectfont}
}

\pgfdeclarelayer{bg}
\pgfsetlayers{bg,main}

\begin{tikzpicture}[node distance=1mm]
    
    \node[fill=orange!80] (node1) at (0,0) {APT-C-36 \faUserSecret};
    \node[right=of node1] {recent analysis};
    \node[fill=green!20] at (1.1, -0.5) {January 15, 2020 \hl{2020-01-15}};
    \node[] (node2) at (-0.4, -1) {From};
    \node[fill=gray!50,right=of node2] (node3) {Lab52 \faBuilding};
    \node[right=of node3] {we have been tracking};
    \node[] at (1.45, -1.5) {through the last months the activity};
    \node[] (node4) at (0, -2) {of the group};
    \node[right=of node4,fill=orange!80] (node5) {APT-C-36 \faUserSecret};
    \node[right=of node5] {. [...]};
    \node[fill=cyan!30] at (1.35, -2.5) {They usually use different types of};
    \node[fill=cyan!30] (techn) at (1.9, -3) {url shorteners in their mailings \hl{\textsc{Technique}}};
    \node[] (node6) at (1.35, -3.5) {[...] Their most popular malware is};
    \node[right=of node6,fill=yellow] (node7) {LimeRAT \faSpider};
    \node[right=of node7]{,};
    \node[] at (2.45,-4) {although many others have been found as indicated};
    \node[] (node8) at (0.1,-4.5) {in the reports.};
    \node[right=of node8,fill=yellow] (node9) {VJWorm \faSpider};
    \node[right=of node9] {has also been seen recently};
    \node[] (node10) at (-0.5,-5) {with};
     \node[fill=blue!20, right=of node10] (node11) {different techniques for exfiltration \hl{\textsc{Tactic}}};
     \node[right=of node11] {.};

     \node[] at (3,-5.6) {\textcolor{orange}{\faUserSecret} \textsc{Group} \hspace{1mm} \textcolor{gray}{\faBuilding} \textsc{Org} \hspace{1mm} \textcolor{yellow}{\faSpider} \textsc{Malware}};

     \node[] at (6, -0.1) {\includegraphics[width=2cm]{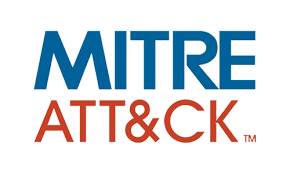}};
     \node[draw,ellipse,text width=1.6cm, align=center,font=\fontsize{7}{7}\selectfont, fill=cyan!30] (kgnode1) at (5.8, -1.4) {T1566/002 Phishing: Spearphishing Link};
     \node[draw,ellipse,text width=1cm, align=center,font=\fontsize{7}{7}\selectfont, fill=orange!80] (kgnode2) at (4.3, 0) {G0099 APT-C-36};
    \node[draw,ellipse,text width=1.2cm, align=center, fill=blue!20,font=\fontsize{7}{7}\selectfont] (kgnode3) at (6.2, -2.7) {TA0010 Exfiltration};

    \draw[dashed, very thick, color=cyan] (kgnode1.south west) .. controls +(-0.3,0) and +(0.3,0) .. (techn.east);
    \draw[dashed, very thick, color=orange] (kgnode2.north) .. controls +(0,0) and +(0,0.4) .. (node1.north);
    \draw[dashed, very thick, color=orange] (kgnode2.south) .. controls +(0.8,-1.9) and +(0.9,0) .. (node5.east);
    \draw[dashed, very thick, color=blue!50] (kgnode3.south) .. controls +(0.5,0) and +(1,0) .. (node11.east);

      \node[ellipse,text width=2mm, align=center,fill=yellow!50] (kgnode6) at (6.9, -.75) {};
       \node[ellipse,text width=3mm, align=center,fill=yellow!50] (kgnode7) at (4.7, -.7) {};
        \node[ellipse,text width=2mm, align=center,fill=orange!30] (kgnode8) at (6.9, -2.15) {};
        \node[ellipse,text width=2mm, align=center,fill=cyan!30] (kgnode9) at (6.6, -4) {};

    \begin{pgfonlayer}{bg}    
        \draw[dotted,color=gray,very thick] (kgnode8) -- (kgnode6);
        \draw[dotted,color=gray,very thick] (kgnode2) -- (kgnode3);
        \draw[dotted,color=gray,very thick] (kgnode8) -- (kgnode1);
        \draw[dotted,color=gray,very thick] (kgnode9) -- (kgnode6);
        \draw[dotted,color=gray,very thick] (kgnode9) -- (7.1,-3.7);
        
    \end{pgfonlayer}

\node[] at (0,-5.7){}; 
\end{tikzpicture}
    
    \caption{\textbf{\corpusName} is a CC-BY-SA-licensed dataset of 120 cyber threat reports annotated with \mitreAttack concepts and WikiData entities. }
    \label{fig:teaser}
\end{figure}
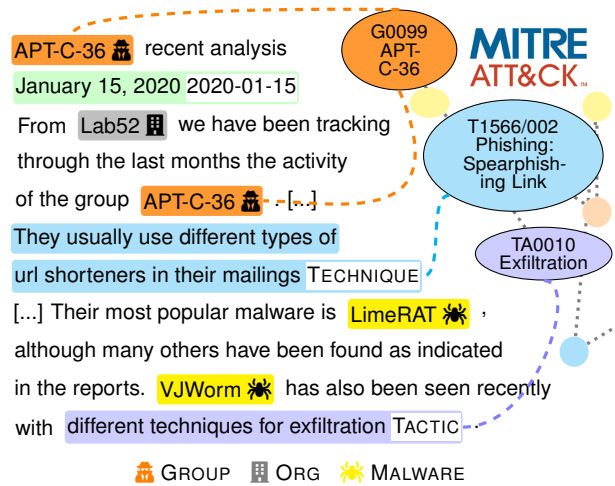

Applying information extraction techniques from natural language processing (NLP) to the domain of CTI is promising, yet understudied. 
MalwareTextDB \citep{lim-etal-2017-malwaretextdb,phandi-etal-2018-semeval} focuses on extracting attributes of malware.
Other datasets \citep[e.g.,][]{Satyapanich_Ferraro_Finin_2020,kim2020automatic} use custom annotation schemas.
In our full-text annotation scenario, we focus on classifying mentions of attack tactics and techniques according to the \mitreAttack taxonomy, a globally accessible database maintained and used by cybersecurity professionals.
We choose this taxonomy rather than a custom schema as it fits directly with the daily work of cybersecurity professionals.
Prior work using \mitreAttack either only annotated entire documents \citep{legoy2020automated} or individual sentences.
Almost all prior datasets are not clearly licensed and hence difficult or even impossible to use without violating copyright.
This aspect has particular relevance in application-driven research.

In this paper, we present \corpusName\footnote{\href{https://github.com/boschresearch/anno-ctr-lrec-coling-2024}{https://github.com/boschresearch/anno-ctr-lrec-coling-2024}}, a new publicly available dataset consisting of 400 CTRs donated by their copyright holders. 
We add annotations targeting information extraction and search tasks, including mentions of locations and organizations linked to Wikipedia, and normalized temporal expressions.
120 of the CTRs have been annotated by a domain expert with cybersecurity-specific concepts, explicitly or implicitly mentioned tactics or techniques from \mitreAttack. 

We propose a set of NLP tasks based on \corpusName.
Transformer-based named entity recognition (NER) models for the general-purpose entities achieve macro-average F1 scores of up to 70\%. 
We find entity disambiguation models \citep{wu-etal-2020-scalable,cao2021genre} fine-tuned on our domain to work well for identifying techniques that occur in a document (micro-F1 of around 65\%).

Our new dataset will enable researchers in NLP and CTI to research, develop, and apply cutting-edge text understanding, search and analysis technology that will provide a very necessary competitive advantage over threat actors.
From an NLP perspective, our contributions are as follows.
\begin{itemize}[nosep,labelindent=0pt]
	\item We provide \corpusName, a new openly-licensed dataset carefully annotated with named entities (NEs) and cybersecurity concepts, and perform a detailed corpus study (\sref{sec:data}).
	\item We propose to model the data from a variety of perspectives using neural sequence tagging, text classification and entity linking models (\sref{sec:modeling}), and provide experimental results for state-of-the-art baselines (\sref{sec:experiments}).
\end{itemize}

\section{Related Work}
\label{sec:relwork}
In this section, we give a brief overview of NLP work and datasets in the cybersecurity domain.

\textbf{\mitreAttack}\footnote{\href{https://attack.mitre.org}{https://attack.mitre.org}, \href{https://github.com/mitre/cti}{https://github.com/mitre/cti}} is a hierarchical knowledge base (KB) of cyber adversary tactics and techniques compiled based on real-world observations.
It is designed to help with managing cyber threat risks.
The KB is regularly updated and released under a license permitting research, development, and commercial use.
At the time of writing, the Enterprise part of the taxonomony, which we are using in this work, consists of 14 tactics and 193 techniques at the top level and 401 subtechniques.
Each technique or tactic comes with a textual description as illustrated in \fref{fig:mitre-technique-example}, as well as several references to CTRs or technical descriptions.
\citet{legoy2020automated} crawl the latter to create a dataset (rcATT) annotated with tactics and techniques at document level.
The CTRs come from many different sources, hence, licensing is unclear.

The Threat Report ATT\&CK Mapper (TRAM) is an open-source platform including a web application aiming to advance research into automating the mapping of CTRs to \mitreAttack.\footnote{\href{https://github.com/center-for-threat-informed-defense/tram}{https://github.com/center-for-threat-informed-defense/tram}}
The TRAM dataset consists of sentences annotated with multi-label \mitreAttack techniques.
Our new dataset and models aim, i.a., to improve the functionality of this open-source endeavor.

\begin{figure}[t]
\footnotesize
\framebox{
\begin{minipage}{0.46\textwidth}
\begin{description}[leftmargin=2mm]
\setlength\itemsep{0.1em}
\item[T1606: Forge Web Credentials] Adversaries may forge credential materials that can be used to gain access to web applications or Internet services. [...]
\begin{description}[leftmargin=2mm]
\setlength\itemsep{0.1em}
\item[T1606.001 Web Cookies] Adversaries may forge web cookies that can be used to gain access to web applications or Internet services. [...]
\item[T1606.002 SAML Tokens] An adversary may forge SAML tokens with any permissions claims and lifetimes if they possess a valid SAML token-signing certificate. [...]
\end{description}
\end{description}
\end{minipage}
}
\caption{\mitreAttack (sub)techniques. }
\label{fig:mitre-technique-example}
\end{figure}

\tref{tab:datasets-overview} gives an overview of manually labeled \textbf{NLP datasets in the cybersecurity domain}.
MalwareTextDB \citep{lim-etal-2017-malwaretextdb,phandi-etal-2018-semeval}
contains reports on hacker groups annotated with the 444 attributes of MAEC\footnote{\href{https://maecproject.github.io/}{https://maecproject.github.io/}} (Malware Attribute Enumeration and Characterization).
The reports are taken from APTnotes\footnote{\href{https://github.com/aptnotes}{https://github.com/aptnotes}} which are publicly available but have unclear licensing.
\citet{hanks2022recognizing} crawl CTI blog posts from the web and conduct a small annotation study for cybersecurity-specific NEs and linking them to Wikipedia.
CASIE \cite{Satyapanich_Ferraro_Finin_2020} and CySecED \cite{man-duc-trong-etal-2020-introducing}
are annotated with cybersecurity event types and semantic arguments.
\citet{kim2020automatic} annotate CTI-Reports with 20 NE types.
\citet{bayer2022multilevel} create a dataset of 3000 tweets annotated for whether they mention a cyber attack.

\begin{table*}[t]
    \centering
    \footnotesize
    \setlength\tabcolsep{3.8pt}
    \scalebox{0.9}{
    \begin{tabular}{lrrrll}
    \toprule
    \textbf{Dataset Name \& Reference} & \textbf{\#Docs.} & \textbf{\#Sents.} & \textbf{\#Annots.} & \textbf{Labels} & \textbf{License}\\
    \midrule
       \href{https://github.com/statnlp-research/statnlp-datasets}{MalwareTextDB-v1} \citep{lim-etal-2017-malwaretextdb} & 39 & 2080 & 7102 & MAEC NE mentions & no license\\
		MalwareTextDB-v2 \citep{phandi-etal-2018-semeval} & 85 & 12,918 & 8054 & MAEC NE mentions & no license\\
		\href{https://github.com/nlpai-lab/CTI-reports-dataset}{CTI-Reports} \citep{kim2020automatic} & 160 & 13,750 & 15,720 & custom NE mentions & no license\\ 
   \href{https://github.com/vlegoy/rcatt}{rcATT} \citep{legoy2020automated} & 1490 & 185,000 & 1490 & \mitreAttack at doc. level & no license\\ 
   \href{https://github.com/UMBC-Onramp/CyEnts-Cyber-Blog-Dataset}{CyEnts} \citep{hanks2022recognizing} & 380 & 1339 & 781 & custom NE mentions & no license \\
  		CySecED \citep{man-duc-trong-etal-2020-introducing} & 292 & $\tilde{}$7300 & 8014 events & 30 event types (w/o arguments) & not available\\
		\href{https://github.com/Ebiquity/CASIE}{CASIE} \citep{Satyapanich_Ferraro_Finin_2020} & 1000 & $\tilde{}$17,000 & 8470 events & 5 event types + arguments & no license\\
    \href{https://github.com/center-for-threat-informed-defense/tram}{TRAM} & -- & 2298 & 3772 & \mitreAttack at sent. level & Apache 2.0\\
    \corpusName (ours) & \numAnnotatedReports & \numSentencesCysecPart & \cysecSpecificAnnotations & NE, \mitreAttack, TIMEX & CC-BY-SA 4.0\\
    \bottomrule
    \end{tabular}
    }
    \caption{Overview of manually labeled \textbf{cybersecurity NLP datasets}. 
    }
    \label{tab:datasets-overview}
\end{table*}

\textbf{NER in the cybersecurity domain} has been modeled using maximum entropy models with n-gram features \citep{bridges2013automatic}, and using tf.idf-based features and word2vec \citep{mikolov2013word2vec} to train a Linear SVM \citep{CortesV95} along with a manually designed confidence propagation procedure \citep{legoy2020automated}.
In the neural age, recurrent neural networks and convolutional neural networks with learned bag-of-characters embeddings and a CRF layer have been used for the task \citep{gasmi2019ner,kim2020automatic,simran2020deep}.
\citet{sarhan2021open} use XLM-R \citep{conneau-etal-2020-unsupervised}.
Further related work aims at constructing or enhancing cybersecurity KBs from text \citep{sarhan2021open,sanagavarapu2021ontoenricher}.

\section{\corpusName Dataset}
\label{sec:data}
In this section, we describe our \corpusName dataset.

\subsection{Source of Texts and Preprocessing}
\corpusName consists of 400 CTRs annotated with general-world entities.
Out of these, 120 reports are also annotated with cybersecurity categories (see \tref{tab:basic_corpus_stats}).
All CTRs have been obtained from the blogs of commercial CTI vendors,\footnote{Intel471, Lab52 (the threat intelligence division of S2 Grupo), Proofpoint, QuoIntelligence, and ZScaler.} who have agreed to their re-publication under CC-BY-SA 4.0.
The blog entries were published between March 2013 and February 2022.
Annotation is performed using the web-based annotation system INCEpTION \citep{klie-etal-2018-inception}.

\textbf{CTI Vendors.}
\label{sec:appendix-vendors}
The reports have been kindly donated by Intel471\footnote{\href{https://intel471.com/blog}{https://intel471.com/blog}}, Lab52\footnote{\href{https://lab52.io/blog/}{https://lab52.io/blog/}} (the threat intelligence division of S2 Grupo\footnote{\href{https://s2grupo.es/}{https://s2grupo.es/}}), Proofpoint\footnote{\href{https://www.proofpoint.com/us/blog}{https://www.proofpoint.com/us/blog}}, QuoIntelligence\footnote{\href{https://quointelligence.eu/blog}{https://quointelligence.eu/blog}}, and ZScaler\footnote{\href{https://www.zscaler.com/blogs/security-research}{https://www.zscaler.com/blogs/security-research}}.

\textbf{Preprocessing.}
First, we retrieve the texts.\footnote{\href{https://github.com/psf/requests}{https://github.com/psf/requests}}
Because of the numerous URLs, code and image references occurring within the texts, we convert the articles into a format similar to Markdown using BeautifulSoupand Markdownify.\footnote{\href{https://www.crummy.com/software/BeautifulSoup/}{https://www.crummy.com/software/BeautifulSoup/}}\textsuperscript{,}\footnote{\href{https://github.com/matthewwithanm/python-markdownify}{github.com/matthewwithanm/python-markdownify}}
Off-the-shelf sentence segmenters do not perform well on texts that contain many links or code snippets, hence, we use a custom regular expression tokenizer for sentence segmentation and correct sentence boundaries manually.

\subsection{Annotation Scheme}
\label{sec:annotation-scheme}
We annotate the reports in our dataset with the following \textbf{General Named Entity} (\textbf{GNE}) types.

\begin{description}[nosep]
    \item[ORG:] Organisations including companies.
    \item[LOC:] Locations, e.g., \textit{California}, \textit{China}.
    \item[SECTOR:] Industry sectors, e.g., \textit{finance}, \textit{defense}.
    \item[TIMEX:] Time expressions for dates normalized following TimeML \citep{sauri2006timeml}, e.g., \textit{July this year} $\rightarrow$ \texttt{2022-07}. 
    \item[CodeSnippet:] Code snippets and command line interface commands.
\end{description}\vspace{2mm} 

\noindent We annotate mentions of \textbf{cybersecurity-specific NEs} (\textbf{CyNE}) as follows.
The term \textit{software} refers to custom or commercial code, operating system utilities, open-source software, or other tools.
    
\begin{description}[nosep]
    \item[GROUP:] Mentions of Advanced Persistant Threats, i.e., hacker groups, e.g., \textit{Fancy Bear}, \textit{Leviathan}, or \textit{APT 40}.
    \item[MALWARE:] Software that has been written specifically for malicious purposes, e.g., \textit{Terdot}.
    \item[TOOL:] Software not written for a malicious purpose but used with a malicious intent in a given context, e.g., \enquote{a malicious \textit{Microsoft Excel} document builder.}
    \item[CONCEPT (CON):] More general concepts relevant to the cybersecurity that can be linked to Wikipedia (e.g., \textit{malware}, \textit{threat actors}), and non-malicious software that is not used maliciously in a context, e.g., \textit{Kaseya VSA}.
    \item[TACTIC:] We annotate mentions of tactics as defined by \mitreAttack
    they capture the adversary's tactical goal (e.g., obtaining credential access), their reason for performing an action.
    \item[TECHNIQUE:] We mark spans that refer to techniques.
    \mitreAttack defines them as follows:
    Techniques represent \textit{how} an adversary achieves a tactical goal by performing an action. 
\end{description}\vspace*{4mm}

\noindent
For TACTIC and TECHNIQUE mentions, we indicate whether the concept is \textit{explicitly} or \textit{implicitly} mentioned.
An explicit mention of a TECHNIQUE means that the descriptive name of the technique is mentioned more or less literally or with a synonym, e.g., as in \textit{send phishing e-mails with malicious attachments} $\rightarrow$ T0865 (Spearphishing Attachment).
Implicit mentions require more inference on the reader's part, as in \textit{Emotet bots reach out to their controllers and received commands to download and execute Trickbot on victim machines.} $\rightarrow$ T1105 (Ingress Tool Transfer).
Explicit mentions are usually short phrases, while implicit phrases may be any part of the text up to a sentence (as also illustrated by \fref{fig:teaser}).

\paragraph{Entity Disambiguation.}
ORG, LOC and CON mentions are linked to Wikipedia pages\footnote{\href{https://en.wikipedia.org/}{hhttps://en.wikipedia.org//}},
GROUP, TACTIC, and TECHNIQUE to \mitreAttack. 

\subsection{Corpus Statistics}
\tref{tab:basic_corpus_stats} shows that the CTRs from the various vendors differ in their average number of sentences, but that sentence lengths are roughly comparable.
\tref{tab:corpus_stats_ne_counts} lists the number of NE mention annotations.
Most of them have been annotated with valid links to Wikipedia or \mitreAttack (exact counts are given in \tref{tab:corpus_stats_links_counts}).
The distribution of techniques and tactics both have a long tail (details in \aref{sec:appendix-corpus-stats}). 
In total, the 120 documents annotated with both layers contain \cysecSpecificAnnotations annotations, the full dataset contains \totalAnnotations annotated mentions.
Hence, as can be seen in \tref{tab:datasets-overview}, \corpusName is at least comparable in size to the frequently used MalwareTextDB v2 \citep{phandi-etal-2018-semeval}.

\begin{table}[t]
    \centering
    \footnotesize
    \setlength\tabcolsep{4pt}
    \scalebox{0.9}{
    \begin{tabular}{lrrrr}
    \toprule
    & \textbf{docs.} & \textbf{sent.} & \textbf{sent/doc} & \textbf{tok./sent}\\
    \midrule
         Intel471 & 30 & 1907 & 63.6$\pm$55.3& 22.3\\
         Lab52 & 23 & 1665 & 72.4$\pm$55.3 & 14.1 \\
         ProofPoint & 28 & 2305 & 82.3$\pm$43.4 & 21.1 \\
         QuoIntelligence & 12 & 1541 & 128.4$\pm$60.2 & 22.4 \\
         ZScaler & 27 & 4761 & 176.3$\pm$107.0 & 21.6 \\
         \midrule
         total & 120 & 12,179 & 101.5$\pm$79.1 & 22.0\\
    \bottomrule
    \end{tabular}
    }
	\caption{\textbf{Corpus statistics} for \corpusName: sentence and token counts (for cyber-security-specific part).} 
    \label{tab:basic_corpus_stats}
\end{table}

\subsection{Annotation Process and Agreement}
Annotation of the general layer (TIMEX, ORG, LOC, SECTOR, and CodeSnippet) was performed by a team of two annotators with an engineering background
who participated in an extensive training phase.
The cybersecurity-specific annotations (all others) were created by a graduate student of media informatics who had previously interned at a cybersecurity group and who hence possesses special domain knowledge.
The annotator was involved in the design of the annotation scheme.
For the \textbf{agreement analysis}, we select 9 documents with a total of 416 sentences: one by QuoIntelligence and two of each other vendor.

\begin{table}[t]
    \centering
    \footnotesize
    \setlength\tabcolsep{4pt}
    \scalebox{0.95}{
\begin{tabular}{lrrrrr|r}
\toprule
\textbf{NE type} &   \rotatebox{90}{Intel471} &  \rotatebox{90}{Lab52} &  \rotatebox{90}{Proofpoint} &  \rotatebox{90}{QuoIntell.} &  \rotatebox{90}{ZScaler} & \rotatebox{90}{total} \\
\midrule
\multicolumn{7}{l}{\textit{Cyber-security specific + general layer (120 docs.)}} \\
\textit{\# docs.} & \textit{30} & \textit{23} & \textit{28} & \textit{12} & \textit{27} & \textit{120}\\
\midrule
TIMEX          &       294 &    155 &         358 &                164 &      155 &  1126 \\
CodeSnippet &        18 &     10 &          32 &                  0 &      255 &   315 \\
LOC         &       220 &    503 &         172 &                174 &       86 &  1155 \\
ORG         &       306 &    377 &         508 &                124 &      263 &  1578 \\
SECTOR      &       132 &     89 &         137 &                131 &       63 &   552 \\
CON         &       236 &     66 &         261 &                185 &      820 &  1568 \\
TOOL        &        24 &     25 &          34 &                 61 &       76 &   220 \\
MALWARE     &       345 &    100 &         474 &                 93 &      657 &  1669 \\
GROUP       &        93 &    102 &         230 &                222 &       66 &   713 \\
TECHN. (expl.)      &       260 &    176 &         229 &                121 &      312 &  1098 \\
TECHN. (impl.) &       301 &    229 &         321 &                198 &     1243 &  2292 \\
TACTIC (expl.)        &        60 &     55 &         219 &                 91 &      271 &   696 \\
TACTIC (impl.)   &        65 &     22 &          19 &                 14 &      142 &   262 \\
\midrule
\multicolumn{7}{l}{\textit{General layer (280 additional docs.)}}\\
\textit{\# docs.} &  \textit{4} & \textit{5} & \textit{73} & \textit{2} & \textit{196} & \textit{280}\\
\midrule
TIMEX          &        61 &     62 &         848 &                 88 &      731 &  1790 \\
CodeSnippet &        32 &      5 &         141 &                  0 &     1324 &  1502 \\
LOC         &        52 &    196 &         732 &                  4 &      271 &  1255 \\
ORG         &        41 &    134 &        1281 &                 27 &     1075 &  2558 \\
SECTOR      &        13 &     36 &         230 &                  8 &      219 &   506 \\
\bottomrule
\end{tabular}
}
    \caption{\textbf{Named entity} annotations in \corpusName.} 
    \label{tab:corpus_stats_ne_counts}
\end{table}

\begin{table}[t]
    \centering
    \footnotesize
    \setlength\tabcolsep{4pt}
    \scalebox{0.95}{
\begin{tabular}{lrrrrr|r}
\toprule
\textbf{NE type} &   \rotatebox{90}{Intel471} &  \rotatebox{90}{Lab52} &  \rotatebox{90}{Proofpoint} &  \rotatebox{90}{QuoIntell.} &  \rotatebox{90}{ZScaler} & \rotatebox{90}{total} \\
\midrule
\multicolumn{7}{l}{\textit{Cyber-security specific + general layer (120 docs.)}} \\
LOC       &       220 &    508 &         172 &                174 &       86 &   1160 \\
ORG       &       200 &    363 &         486 &                124 &      263 &   1436 \\
CON       &       220 &     66 &         261 &                178 &      814 &   1539 \\
TOOL      &        24 &     25 &          34 &                 61 &       75 &    219 \\
MALWARE   &       345 &    100 &         474 &                 93 &      657 &   1669 \\
GROUP     &        93 &    103 &         230 &                224 &       66 &    716 \\
TECHNIQUE &       560 &    404 &         550 &                319 &     1542 &   3375 \\
TACTIC    &       125 &     77 &         238 &                105 &      413 &    958 \\
\midrule
\multicolumn{7}{l}{\textit{General layer (280 additional docs.)}}\\
LOC    &        52 &    201 &         732 &                  4 &      271 &   1260 \\
ORG    &        22 &    133 &        1261 &                 27 &     1075 &   2518 \\
\bottomrule
\end{tabular}
}
    \caption{\textbf{Disambiguated entity mentions} in \corpusName with links to \mitreAttack or WikiData.} 
    \label{tab:corpus_stats_links_counts}
\end{table}

\textbf{General annotations.}
The nine documents of the agreement study were marked by an additional annotator, who had not been involved in the first phase, but had already received extensive training on the almost same annotation task in a different domain.
When comparing TIMEX annotations, we find 41 exact matches.
Six additional cases are annotator lapses, i.e., trivial mistakes. 
In one other case, one annotator did not include the \textit{late} into the span of \textit{late October}.
There are 18 LOC annotations with an observed agreement of 100\% on spans.
For 17 of these, the annotators agree on the Wikipedia link. 
For ORG, 73 and 81 instances are marked by the two annotators, respectively, amounting to precision and recall scores of 78.1 and 70.4\% for exact matches.
When applying relaxed matching (containment), the scores go up to 89.0 and 80.2\%.
Out of 57 exact matches and 8 relaxed matches, we found only 3 Wikipedia links to differ. 
The two annotators mark only 10 and 20 SECTOR mentions, respectively, with a relaxed agreement of 70.0 and 35.0\%.

\textbf{Cybersecurity-specific annotations.}
This type of annotation requires specialized cybersecurity knowledge and a high familiarity with the \mitreAttack taxonomy.
The entire dataset has been marked by single domain expert annotator.
In order to provide a rough estimate for the difficulty of the annotation task, we ask another cybersecurity professional with a degree in media informatics and 5 years of professional experience to label our data.
As their time constraints did not allow an extended training phase, the agreement presented here in all likelihood underestimates the degree of agreement that is achievable with more training.
Precision and recall for identifying entity mentions amount to F1-scores of 31\% for CON, 52\% for GROUP, and 67\% for MALWARE.
The second annotator did not mark any TOOL annotations.
When comparing the sets of techniques found in a document, 
average F1 amounts to 54\%. 
While this study does not constitute a proper agreement study (as we acknowledge would be desirable), it still demonstrates that annotations are systematic.
An exception may be CON, where the main annotator considered a different set of concepts as relevant.
Yet, this set seems to be consistent as in our experiments, the tag can be learned well (F1 68\%).

\label{sec:models}
\label{sec:modeling}
\section{Task Definitions and Modeling}
\label{sec:modeling}

In this section, we define several NLP tasks based on \corpusName and describe neural models that we propose as strong baselines for solving them.

\subsection{Named Entity Recognition}
\label{sec:modeling-ner}
The first processing step consists of detecting entity mentions and tagging them with the NE types listed in \sref{sec:annotation-scheme}.
We represent label sequences using the BIO scheme.
First, the model processes the input sequence with a pre-trained transformer model, applies a linear classification layer to the transformer output to compute the logits for each potential NE label, and predicts the label corresponding to the maximum logit for each first wordpiece token of each \enquote{real} token. 

We compare the following pre-trained transformer models: BERT \citep{devlin-etal-2019-bert}, which has been trained on Google Books \citep{zhu2015googlebooks} and the English Wikipedia, SciBERT \citep{beltagy-etal-2019-scibert}, a BERT-style model trained on scientific text, and CodeBERT \citep{feng-etal-2020-codebert}, which has been trained on programming code.
We also use RoBERTA \citep{liu2019roberta}, which is trained on Google Books, Wikipedia, and news articles, and which uses an optimized training procedure compared to BERT.

As a recent baseline for NER in the cybersecurity domain, \citet{gasmi2019ner} and \citet{kim2020automatic} use a BiLSTM-CRF \citep{lample-etal-2016-neural} with GloVe \citep{pennington-etal-2014-glove} embeddings, essentially following the architecture of \citet{huang2015ner}. We compare to a reimplementation of the model by \citet{kim2020automatic} in our experiments.

\subsection{Temporal Tagging}
We compare two methods for temporal expression extraction and normalization. 
First, we use the rule-based system Heideltime \citep{strotgen-gertz-2010-heideltime} limited to DATE rules.
We experiment with adapting HeidelTime to the cybersecurity domain by automatically selecting a subset of rules.
We sequentially remove randomly selected rules and keep only rules that influence the performance positively. We start this procedure with 5 different random seeds and keep the best-performing subset of rules based on the dev set score. 

Second, we use a multilingual neural temporal tagging model \citep{lange2022temporal}.
This model first detects TIMEXes using a neural sequence tagger trained on gold standard corpora of news and Wikipedia, and then, for temporal expression normalization, uses a masked language modeling approach to fill slots in XML-like templates:

{\footnotesize
\texttt{The attack happened <TIMEX type="DATE" value="\textbf{YEAR}-\textbf{MONTH}-\textbf{DAY}">yesterday}\texttt{</TIMEX>}.}

\textbf{\texttt{YEAR}}, \textbf{\texttt{MONTH}}, and \textbf{\texttt{DAY}} are masks, whose values are predicted by the language model.
The normalization model has been trained on a multilingual weakly-supervised dataset created using HeidelTime.
In our experiments, we substitute the sequence tagger for TIMEX detection with our cybersecurity-specific NER model  (\sref{sec:modeling-ner}).

\subsection{Entity Disambiguation}
\label{sec:modeling-entity-disambiguation}
NE mentions in \corpusName are, depending on their type, linked to Wikipedia 
or \mitreAttack.
We restrict the search space for the linking models depending on the extracted NE type.
LOC/ORG/CON are linked against Wikipedia.
MALWARE and TOOL are linked to the \verb|/software/| subtree of \mitreAttack. Accordingly, we map GROUPS to  \verb|/groups/|, TACTIC to \verb|/tactics/| and TECHNIQUES to \verb|/techniques/| only. 
In this section, we describe several models that we use to identify the KB entry disambiguating the mention.

\textbf{BLINK} \citep{wu-etal-2020-scalable} uses a bi-encoder model to compute embeddings for all entries of the KB and for the input sentence with the mention to be disambiguated.
A set of candidates is selected based on the similarity of the target sentence encoding to the KB encodings.
In a second step, a cross-encoder computes a final ranking score for the concatenation of the text strings representing the entity in the KB and the input sentence. 
In our experiments, we found that in our setting, the cross-encoder does not provide any benefit and hence only use the bi-encoder part of the model.
BLINK has been trained on 9M examples of document-mention-entity triples from Wikipedia. 

\textbf{GENRE} \citep{cao2021genre} is a constrained language model based on BART \citep{lewis-etal-2020-bart}.
Given an input sentence with an entity to be linked (marked using special tags), the model is asked to generate the most likely KB entry.
The search space is constrained by the entries of the target KB, e.g., all titles of Wikipedia pages for the general-domain linking, or the titles or \mitreAttack entries for mentions of cybersecurity entities.
GENRE has first been trained on the same Wikipedia data as BLINK and then further fine-tuned on AIDA-CoNLL \citep{hoffart-etal-2011-robust}.

BLINK works out of the box for linking to \mitreAttack as the pre-trained bi-encoder can be applied to any KG.
We use the \mitreAttackShort descriptions (as in \fref
{fig:mitre-technique-example}) to generate entity representations. 
GENRE can also be applied in a zero-shot setting on \mitreAttack as the set of KG entities that constrains decoding is part of the model configuration, which are the \mitreAttackShort titles in our setting. 
In order to compare general-purpose entity disambiguation models and cybersecurity-specific models, we 
fine-tune the models on \corpusName.
In cases where BLINK or GENRE predict a subtechnique, we change the prediction to the parent technique.

\subsection{Sentence-based Tactic and Technique Classification}
\label{sec:modeling-text-classification}
From a practical point, it matters to correctly detect the set of tactics and techniques that occur in a document.
In particular, implicitly mentioned techniques are often marked as long phrases and hence NER models are not well-suited.
We here address the problem using few-shot text classification methods.
In order to comprehensively detect and link explicit and implicit cyber attack tactics and techniques to \mitreAttack, we first detect sentences mentioning a TACTIC or TECHNIQUE, and then classify these sentences into the set of concepts as defined by \mitreAttackShort.

\textbf{Detection.}
First, we train two four-way sentence classifiers using RoBERTA for detecting TACTICs and TECHNIQUEs, respectively.
The classifiers predict whether an input sentence contains an explicit, and implicit, both an explicit and an implicit, or no mention of TECHNIQUE/TACTIC.
This setup performed better or on par with a binary classifier.
The second step takes as an input sentences that are predicted to contain TECHNIQUE/TACTIC mentions and the linking model links the sentence to a node in \mitreAttackShort.

\paragraph{Classification/Disambiguation.}
For classifying a sentence into a set of tactics or techniques, we compare the following models.
\begin{description}[nosep]
    \item[GENRE:] We use the entity linking model described in \sref{sec:modeling-entity-disambiguation}, but do not mark entity mentions in the input sentence.
    \item[TMM:] The Transformer-based Multi Task Model 
 \citep{pujari2021multi} encodes the input sentence using SciBERT, and feeds the CLS embedding into a set of binary classification heads which each predict whether a particular technique occurs in the sentence or not.
    \item[TRAM:] For comparison, we run the model provided by the TRAM project on our data. The model consists of an ensemble of a logistic regression and a Naive Bayes model using n-gram features and has been trained on the TRAM dataset for technique detection and linking. 
\end{description}

\label{sec:experiments}
\section{Experiments}
\label{sec:experiments}
This section describes our experimental results on our \corpusName corpus for the NLP tasks and models introduced in the previous sections.

\subsection{Settings}
For our experiments, we split the \corpusName corpus into three parts (train/dev/test) with 60/15/25\% of the documents, respectively. 
We perform a temporal split ensuring that documents from each vendor end up in each split.
The temporal split simulates a real-world scenario as older documents are used for training and more recent documents are processed during inference.
The dev set is used for model picking. 
Results are reported on the test set.
For the NER experiments, we train 5 models with different random seeds and report the average scores and standard deviation of all runs.

\textbf{Hyperparameters.}
We did not perform a hyperparameter search, as we used the suggested default values, i.e., the Huggingface training script for NER,\footnote{\href{https://github.com/huggingface/transformers/blob/main/examples/pytorch/token-classification/run\_ner.py}{https://github.com/huggingface/.../run\_ner.py}} and the model-specific scripts for BLINK and GENRE from the respective repositories.
For GENRE, we set the number of beams and output sequences to 10 during decoding to match the number of candidates of BLINK.

\subsection{NER Results}
\tref{tab:ner-results} contains the results for our NER experiments.
In this experiment, we only use explicit mentions of TECHNIQUE and TACTIC, as we found in preliminary experiments that the NER models do not work well for implicit mentions.
The BiLSTM-CRF baseline \citep{kim2020automatic} performs between 12 and 15 F1 points worse than the transformers. 
Differences between the transformer models are smaller.
The RoBERTa models perform best for general and cybersecurity entities with 2-3 points improvements compared to the other BERT models.
A possible explanation is that the pretraining data of RoBERTa is closer to \corpusName.
For example, in contrast to the pretraining data of the other models, it includes news articles, which are comparable in structure to CTRs. 
Moreover, training all models on the extended corpus with general entities ($^{+X}$) consistently improves the performance by~2~points.   

\begin{table}[t]
\footnotesize
\centering
\setlength\tabcolsep{4pt}
\scalebox{0.95}{
\begin{tabular}{lccc} \toprule
Model      & GNE & GNE$^{+X}$ & CyNE  \\ \midrule
BiLSTM-CRF {\small (\citeauthor{kim2020automatic})} & 62.2\sd{0.6} & 65.2\sd{1.1} & 43.7\sd{1.6} \\
BERT       & 77.2\sd{1.7} & 80.4\sd{1.3} & 51.0\sd{2.3} \\
SciBERT    & 76.3\sd{1.8} & 80.4\sd{1.4} & 53.0\sd{2.3} \\
CodeBERT   & 75.7\sd{2.0} & 80.3\sd{1.9} & 53.7\sd{3.5} \\
RoBERTa    & \textbf{80.2\sd{1.6}} & \textbf{82.3\sd{1.2}} & \textbf{57.8\sd{1.9}} \\
\bottomrule
\end{tabular}
}
\caption{\textbf{NER results}. 
Metric: average Micro F1 of 5 runs for general-world named entities (GNE) and \textit{explicit} cybersecurity entities (CyNE) trained on 120 documents. GNE$^{+X}$: training includes data from 280 additional documents.}
\label{tab:ner-results}
\end{table}

\subsection{Temporal Tagging Results}
The results for temporal tagging are given in \tref{tab:temporal-results}. We use the TempEval3 evaluation script \citep{uzzaman-etal-2013-semeval} and report relaxed/strict F1 and value F1 for the extraction and normalization of temporal expressions, respectively.

Both out-of-the-box models, HeidelTime and the neural model of \citet{lange2022temporal}, have rather low scores for the extraction step.
Thus, we experimented with domain adaptation methods and find that both can be greatly improved.
First, the HeidelTime model benefits from the reduced set of rules for the cybersecurity domain (+12.0/10.5/9.2 pp.), which results in a decrease of false positive matches.
These are often caused by differences in the annotation schemes, e.g., imprecise expressions like \textit{now} or \textit{soon} are not annotated in \corpusName.
Second, the neural model can be improved by substituting the extraction component with our domain-specific cybersecurity NER model (+18.5/11.0 pp.). 
As a result, the normalization relying on the extracted expressions performs better as well (+10.6 pp.).

\begin{table}[t]
\footnotesize
\centering
\setlength\tabcolsep{4pt}
\scalebox{0.95}{
\begin{tabular}{llll} \toprule
      & \multicolumn{2}{c}{\textit{Extraction}} & \textit{Norm.} \\
Model & Strict  & Relaxed & Value \\ \midrule
HeidelTime & 57.5 & 69.3 & 69.3 \\
 + optimized rules               & 69.5 & 79.8 & 78.5 \\
NER+MLM {\small (\citeauthor{lange2022temporal})} & 68.7 & 83.2 & 81.6 \\
 + cysec. NER model              & 84.3\sd{1.7} & 93.4\sd{0.8} & 89.2\sd{0.8} \\
 + cysec. NER model$^{+X}$       & \textbf{87.2\sd{0.7}} & \textbf{94.2\sd{0.4}} & \textbf{92.2\sd{0.3}} \\ \bottomrule
\end{tabular}
}
\caption{\textbf{Temporal tagging} results. 
Metrics: F1 sores for the extraction and normalization.}
\label{tab:temporal-results}
\end{table}

\subsection{Entity and Concept Disambiguation}
\label{sub:entity-linking}
We report the results for entity disambiguation when linking against Wikipedia in \tref{tab:linking-wiki-results}.
While precision is higher for the top-1 predictions of GENRE, it suffers from lower recall compared to BLINK for ORG and CON entities.
This means that likely, GENRE prunes away good candidates too early.
The scores for ORG are lower compared to the other two types, indicating that selecting the correct entity is harder for ORGs, often due to the presence of several likely candidates.

\begin{table}[t]
\footnotesize
\centering
\setlength\tabcolsep{4pt}
\scalebox{0.95}{
\begin{tabular}{lcc|cc|cc} \toprule
 & \multicolumn{2}{c|}{ORG} & \multicolumn{2}{c|}{LOC} & \multicolumn{2}{c}{CON} \\
Model & Acc. & R@10 & Acc. & R@10 & Acc. & R@10 \\ \midrule
BLINK & 50.4 & \textbf{66.4} & 85.7 & 97.8 & 64.5 & \textbf{87.3} \\
GENRE & \textbf{57.2} & 60.0 & \textbf{93.7} & \textbf{97.9} & \textbf{64.7} & 77.9 \\ \bottomrule
\end{tabular}
}
\caption{\textbf{Entity disambiguation} results on gold-standard general entities for linking against \textbf{Wikipedia}. Metrics: accuracy (R@1) / recall@10. }
\label{tab:linking-wiki-results}
\end{table}

\begin{table*}[htb]
\footnotesize
\centering
\setlength\tabcolsep{6pt}
\begin{tabular}{lcc|cc|cc|cc|cc} \toprule
 & \multicolumn{2}{c|}{GROUP} & \multicolumn{2}{c|}{MALWARE} & \multicolumn{2}{c|}{TOOL} & \multicolumn{2}{c|}{TACTIC} & \multicolumn{2}{c}{TECHNIQUE} \\
Model & Acc. & R@10 & Acc. & R@10 & Acc. & R@10 & Acc. & R@10 & Acc. & R@10 \\ \midrule
BLINK (zero-shot) & 82.6 & 85.5 & 91.0 & 96.7 & \textbf{100.0}  & \textbf{100.0}  & 57.7 & 82.0 & 11.6 & 29.2 \\
+ finetuned      & 82.2 & 87.8 & 89.1 & 96.9 & 98.1 & \textbf{100.0}  & \textbf{85.1} & 95.7 & 45.1 & 68.6 \\
GENRE (zero-shot) & \textbf{86.4} & \textbf{89.2} & \textbf{95.0} & \textbf{97.1} & 88.5 & 88.5 & 66.7 & 75.7 & 14.6 & 19.8 \\
+ finetuned      & 11.5 & 28.8 & 64.8 & 64.8 & 9.9 & 27.3 & 64.6 & 80.6 & 39.7 & 65.0 \\
retrained & 49.8 & 89.2 & 64.4 & 97.0 & 46.2 & \textbf{100.0} & 83.5 & \textbf{96.5} & \textbf{66.9} & \textbf{87.8} \\ \midrule
\end{tabular}
\caption{\textbf{Entity disambiguation} results on gold-standard cybersecurity entities for linking against \textbf{\mitreAttack}. Metrics: accuracy (R@1) / recall@10.
Explicit and implicit TACTICs and TECHNIQUEs.
}
\label{tab:linking-mitre-results}
\end{table*}

Results for linking against \mitreAttack are shown in in \tref{tab:linking-mitre-results}.
In our zero-shot experiments, we use the standard BLINK and GENRE models trained on Wikipedia (+ AIDA-CoNLL) and evaluate
them for linking to \mitreAttack.
We also experiment with fine-tuning the models on our corpus and we retrain a GENRE model from an initial BART checkpoint without any previous entity disambiguation training.
We find that the zero-shot models perform already reasonably well on GROUP, MALWARE, and TOOL.
Fine-tuning on \corpusName for these entity types mostly reduces the performance, which may be caused by catastrophic forgetting of the original training or overfitting. 

By contrast, fine-tuning on \corpusName is essential for correctly linking tactics and techniques.
In particular, the retrained GENRE model outperforms all other models for techniques by a large margin. 
This task is quite different from standard entity disambiguation, as the textual surface form often notably differs from the KB concept's title in contrast to when linking against Wikipedia or to linking GROUP/MALWARE/TOOL.
The experiments in \tref{tab:linking-mitre-results} use gold standard entity spans and types.

\subsection{Text-Classification-based Tactic and Technique Disambiguation Results}
\label{sub:entity-linking-sentence}
We evaluate the identification of tactics and techniques mentioned in a document in an end-to-end setting, 
as proposed in \sref{sec:modeling-text-classification}.
As a first step, we train a RoBERTA-based text classification model to identify sentences containing a TECHNIQUE or TACTIC mention.
The model achieves F1 scores of 77.0\% and 59.4\% for the two labels, respectively.

The sentences that are predicted to mention a tactic or technique are then classified with regard to which \mitreAttack concepts they mention.
We retrain several GENRE models, as this model has shown the most promising results for these classes in our previous experiments.
Moreover, we augment the training data with an additional 7972 sentences taken from the technique/tactic descriptions as found in \mitreAttackShort, labeled with the corresponding tactic/technique.
For techniques, we also add the TRAM dataset to the training.
As a baseline for comparison, we train the GENRE model with a negative class for sentences that do not contain a TACTIC or TECHNIQUE.
This model receives all sentences as input.
In order to estimate the impact of the first sentence classification step, we compare to using the sentences containing a TECHNIQUE or TACTIC annotation in the gold standard.

\tref{tab:doc-level-techniques} shows the results in terms of the average F1 score of detecting the set of 
\mitreAttackShort techniques/tactics mentioned in a document.
We find that our domain-specialized models perform a lot better than the zero-shot model and the baselines.
In particular, the models leveraging the extra training data from TRAM or the \mitreAttack descriptions work best.
For techniques, we find that the approach utilizing our text classification model for the sentence filtering performs the best.
In contrast, this method performs worse than the linking model using a negative class for tactics.
As indicated by the high scores when using oracle sentence selection (lower part of the table), this is caused by the worse performance of the text classification model for TACTIC, which has less than a third training examples compared to TECHNIQUE.
We assume that a better text classification model for TACTIC entities will also improve the performance of our GENRE models as suggested by the gold-standard sentence selection results.

\begin{table}[t]
\footnotesize
\centering
\setlength\tabcolsep{4pt}
\scalebox{0.95}{
\begin{tabular}{lcc} \toprule
\textbf{Model}     & \rotatebox{90}{\textbf{Techn.}} & \rotatebox{90}{\textbf{Tactic}} \\ \midrule
TRAM baseline  & 24.8 & -\\
TMM \citep{pujari2021multi} & 35.3 & 36.1 \\
\midrule
\multicolumn{3}{l}{\textit{Sentence-level linking models with negative class}} \\
GENRE (\corpusName)            & 42.5 & 45.8 \\
GENRE (\corpusName+KG desc.) & 45.2 & 47.7 \\
GENRE (\corpusName+TRAM)     & 47.1 & -\\
GENRE (\corpusName+TRAM+KG desc.) & 43.7 & -\\
GENRE (TRAM)                   & 25.3 & -\\
\midrule
\multicolumn{3}{l}{\textit{Sentence-level linking models + text classification}} \\
GENRE (zero-shot)              & 23.4 & 21.9 \\
GENRE (\corpusName)            & 52.9 & \textbf{39.2} \\
GENRE (\corpusName+KG desc.) & \textbf{56.6} & 36.7 \\
GENRE (\corpusName+TRAM)     & 56.0 & -\\
GENRE (\corpusName+TRAM + KG desc.) & 56.5 & -\\
GENRE (TRAM)                   & 25.9 & -\\
\midrule
\multicolumn{3}{l}{\textit{Sentence-level linking models + gold techn. detection}} \\
GENRE (zero-shot)              & 24.8 & 51.7 \\
GENRE (\corpusName)            & 58.9 & \textbf{84.2} \\
GENRE (\corpusName+KG desc.) & \textbf{65.7} & 83.3 \\
GENRE (\corpusName+TRAM)     & 65.4 & -\\
GENRE (\corpusName+TRAM+KG desc.) & 63.7 & -\\
GENRE (TRAM)                  & 27.7 & -\\
\bottomrule 
\end{tabular}
}
\caption{\textbf{Document-level technique and tactic detection} results: F1 scores micro-averaged over documents. }
\label{tab:doc-level-techniques}
\end{table}

\section{Conclusion and Outlook}
\label{sec:conclusion}
In this resource paper, we have described a new large-scale dataset in the cybersecurity domain annotated with general-world NEs and cybersecurity concepts including tactics and techniques.
We have proposed several NLP tasks and provided an extensive set of experimental results using neural transformer models for NER and linking entities and concepts to Wikipedia and \mitreAttack, demonstrating that the corpus is consistently annotated.
Our work lays the foundation for developing cybersecurity-specific NLP models using freely available and permissively licensed data. 

\section*{Ethical Considerations}
The annotators participating in our project were aware of the goal of the annotations and gave their explicit consent to the publication of their annotations.
The annotators were paid considerably above the respective country's minimum wages.

While our work attempts to help fighting cybercrime, it is like most NLP work not exempt from the risk of dual use.

\section*{Limitations}
Our experiments are focused on the \corpusName dataset that we describe in this paper. We could not perform larger-scale multi-task or transfer learning with other datasets due to licensing issues as mentioned in \sref{sec:relwork}. 
The exception for which we could try transfer learning was TRAM. 
While the experiments in \sref{sub:entity-linking-sentence} resemble a real-world evaluation, the linking models in \sref{sub:entity-linking} take the gold-standard entities as inputs, which assumes a perfect extraction model. 
The training of any of our neural models requires a considerable number of computational resources (up to 12 GPU hours), which might not be available for every person/organization.

As knowledge bases are typically continually updated, the recognition and linking models have to deal with unseen classes during real-world inference setups. 
While the corpus itself cannot cover these new concepts without constant updates, our models are adaptable to such changes. 
For the entity recognition task, our annotated data guides the model to identify contexts in which an entity usually appears, and hence a full enumeration of possible values is not even the case at present.
For entity linking, we require a snapshot of the database, from which we extract a list of all concepts, techniques, tactics, etc.~with corresponding textual descriptions. These descriptions are used in the models as targets for decoding, i.e., the model has to output valid entries from the given snapshot. We can exchange or update the knowledge base, such that the model can also output new entities.

\newpage
\section{Bibliographical References}\label{sec:reference}

\bibliographystyle{lrec-coling2024-natbib}
\bibliography{citations,anthology}

\begin{thebibliography}{35}
\expandafter\ifx\csname natexlab\endcsname\relax\def\natexlab#1{#1}\fi

\bibitem[{Bayer et~al.(2022)Bayer, Frey, and Reuter}]{bayer2022multilevel}
Markus Bayer, Tobias Frey, and Christian Reuter. 2022.
\newblock \href {https://doi.org/10.48550/arXiv.2207.11076} {Multi-level fine-tuning, data augmentation, and few-shot learning for specialized cyber threat intelligence}.
\newblock \emph{CoRR}, abs/2207.11076.

\bibitem[{Beltagy et~al.(2019)Beltagy, Lo, and Cohan}]{beltagy-etal-2019-scibert}
Iz~Beltagy, Kyle Lo, and Arman Cohan. 2019.
\newblock \href {https://doi.org/10.18653/v1/D19-1371} {{S}ci{BERT}: A pretrained language model for scientific text}.
\newblock In \emph{Proceedings of the 2019 Conference on Empirical Methods in Natural Language Processing and the 9th International Joint Conference on Natural Language Processing (EMNLP-IJCNLP)}, pages 3615--3620, Hong Kong, China. Association for Computational Linguistics.

\bibitem[{Bridges et~al.(2013)Bridges, Jones, Iannacone, and Goodall}]{bridges2013automatic}
Robert~A. Bridges, Corinne~L. Jones, Michael~D. Iannacone, and John~R. Goodall. 2013.
\newblock \href {http://arxiv.org/abs/1308.4941} {Automatic labeling for entity extraction in cyber security}.
\newblock \emph{CoRR}, abs/1308.4941.

\bibitem[{Cao et~al.(2021)Cao, Izacard, Riedel, and Petroni}]{cao2021genre}
Nicola~De Cao, Gautier Izacard, Sebastian Riedel, and Fabio Petroni. 2021.
\newblock \href {https://openreview.net/forum?id=5k8F6UU39V} {Autoregressive entity retrieval}.
\newblock In \emph{9th International Conference on Learning Representations, {ICLR} 2021, Virtual Event, Austria, May 3-7, 2021}. OpenReview.net.

\bibitem[{Conneau et~al.(2020)Conneau, Khandelwal, Goyal, Chaudhary, Wenzek, Guzm{\'a}n, Grave, Ott, Zettlemoyer, and Stoyanov}]{conneau-etal-2020-unsupervised}
Alexis Conneau, Kartikay Khandelwal, Naman Goyal, Vishrav Chaudhary, Guillaume Wenzek, Francisco Guzm{\'a}n, Edouard Grave, Myle Ott, Luke Zettlemoyer, and Veselin Stoyanov. 2020.
\newblock \href {https://doi.org/10.18653/v1/2020.acl-main.747} {Unsupervised cross-lingual representation learning at scale}.
\newblock In \emph{Proceedings of the 58th Annual Meeting of the Association for Computational Linguistics}, pages 8440--8451, Online. Association for Computational Linguistics.

\bibitem[{Cortes and Vapnik(1995)}]{CortesV95}
Corinna Cortes and Vladimir Vapnik. 1995.
\newblock \href {https://doi.org/10.1007/BF00994018} {Support-vector networks}.
\newblock \emph{Mach. Learn.}, 20(3):273--297.

\bibitem[{Devlin et~al.(2019)Devlin, Chang, Lee, and Toutanova}]{devlin-etal-2019-bert}
Jacob Devlin, Ming-Wei Chang, Kenton Lee, and Kristina Toutanova. 2019.
\newblock \href {https://doi.org/10.18653/v1/N19-1423} {{BERT}: Pre-training of deep bidirectional transformers for language understanding}.
\newblock In \emph{Proceedings of the 2019 Conference of the North {A}merican Chapter of the Association for Computational Linguistics: Human Language Technologies, Volume 1 (Long and Short Papers)}, pages 4171--4186, Minneapolis, Minnesota. Association for Computational Linguistics.

\bibitem[{Feng et~al.(2020)Feng, Guo, Tang, Duan, Feng, Gong, Shou, Qin, Liu, Jiang, and Zhou}]{feng-etal-2020-codebert}
Zhangyin Feng, Daya Guo, Duyu Tang, Nan Duan, Xiaocheng Feng, Ming Gong, Linjun Shou, Bing Qin, Ting Liu, Daxin Jiang, and Ming Zhou. 2020.
\newblock \href {https://doi.org/10.18653/v1/2020.findings-emnlp.139} {{C}ode{BERT}: A pre-trained model for programming and natural languages}.
\newblock In \emph{Findings of the Association for Computational Linguistics: EMNLP 2020}, pages 1536--1547, Online. Association for Computational Linguistics.

\bibitem[{Gasmi et~al.(2019)Gasmi, Laval, and Bouras}]{gasmi2019ner}
Houssem Gasmi, Jannik Laval, and Abdelaziz Bouras. 2019.
\newblock \href {https://doi.org/10.3390/app9193945} {Information extraction of cybersecurity concepts: An lstm approach}.
\newblock \emph{Applied Sciences}, 9(19).

\bibitem[{Hanks et~al.(2022)Hanks, Maiden, Ranade, Finin, Joshi et~al.}]{hanks2022recognizing}
Casey Hanks, Michael Maiden, Priyanka Ranade, Tim Finin, Anupam Joshi, et~al. 2022.
\newblock Recognizing and extracting cybersecurity entities from text.
\newblock In \emph{Workshop on Machine Learning for Cybersecurity, International Conference on Machine Learning}.

\bibitem[{Hoffart et~al.(2011)Hoffart, Yosef, Bordino, F{\"u}rstenau, Pinkal, Spaniol, Taneva, Thater, and Weikum}]{hoffart-etal-2011-robust}
Johannes Hoffart, Mohamed~Amir Yosef, Ilaria Bordino, Hagen F{\"u}rstenau, Manfred Pinkal, Marc Spaniol, Bilyana Taneva, Stefan Thater, and Gerhard Weikum. 2011.
\newblock \href {https://aclanthology.org/D11-1072} {Robust disambiguation of named entities in text}.
\newblock In \emph{Proceedings of the 2011 Conference on Empirical Methods in Natural Language Processing}, pages 782--792, Edinburgh, Scotland, UK. Association for Computational Linguistics.

\bibitem[{Huang et~al.(2015)Huang, Xu, and Yu}]{huang2015ner}
Zhiheng Huang, Wei Xu, and Kai Yu. 2015.
\newblock \href {http://arxiv.org/abs/1508.01991} {Bidirectional {LSTM-CRF} models for sequence tagging}.
\newblock \emph{CoRR}, abs/1508.01991.

\bibitem[{Kim et~al.(2020)Kim, Lee, Jo, and Lim}]{kim2020automatic}
Gyeongmin Kim, Chanhee Lee, Jaechoon Jo, and Heuiseok Lim. 2020.
\newblock \href {https://doi.org/10.1007/s13042-020-01122-6} {{Automatic extraction of named entities of cyber threats using a deep Bi-LSTM-CRF network}}.
\newblock \emph{International Journal of Machine Learning and Cybernetics}, 11(10):2341--2355.
\newblock Funding Information: Funding was provide by Korea Creative Content Agency (Grant No. R2017030045). Publisher Copyright: {\textcopyright} 2020, Springer-Verlag GmbH Germany, part of Springer Nature.

\bibitem[{Klie et~al.(2018)Klie, Bugert, Boullosa, Eckart~de Castilho, and Gurevych}]{klie-etal-2018-inception}
Jan-Christoph Klie, Michael Bugert, Beto Boullosa, Richard Eckart~de Castilho, and Iryna Gurevych. 2018.
\newblock \href {https://aclanthology.org/C18-2002} {The {INCE}p{TION} platform: Machine-assisted and knowledge-oriented interactive annotation}.
\newblock In \emph{Proceedings of the 27th International Conference on Computational Linguistics: System Demonstrations}, pages 5--9, Santa Fe, New Mexico. Association for Computational Linguistics.

\bibitem[{Lample et~al.(2016)Lample, Ballesteros, Subramanian, Kawakami, and Dyer}]{lample-etal-2016-neural}
Guillaume Lample, Miguel Ballesteros, Sandeep Subramanian, Kazuya Kawakami, and Chris Dyer. 2016.
\newblock \href {https://doi.org/10.18653/v1/N16-1030} {Neural architectures for named entity recognition}.
\newblock In \emph{Proceedings of the 2016 Conference of the North {A}merican Chapter of the Association for Computational Linguistics: Human Language Technologies}, pages 260--270, San Diego, California. Association for Computational Linguistics.

\bibitem[{Lange et~al.(2022)Lange, Str{\"{o}}tgen, Adel, and Klakow}]{lange2022temporal}
Lukas Lange, Jannik Str{\"{o}}tgen, Heike Adel, and Dietrich Klakow. 2022.
\newblock \href {https://doi.org/10.48550/arXiv.2205.10399} {Multilingual normalization of temporal expressions with masked language models}.
\newblock \emph{CoRR}, abs/2205.10399.

\bibitem[{Legoy et~al.(2020)Legoy, Caselli, Seifert, and Peter}]{legoy2020automated}
Valentine Legoy, Marco Caselli, Christin Seifert, and Andreas Peter. 2020.
\newblock \href {http://arxiv.org/abs/2004.14322} {Automated retrieval of att{\&}ck tactics and techniques for cyber threat reports}.
\newblock \emph{CoRR}, abs/2004.14322.

\bibitem[{Lewis et~al.(2020)Lewis, Liu, Goyal, Ghazvininejad, Mohamed, Levy, Stoyanov, and Zettlemoyer}]{lewis-etal-2020-bart}
Mike Lewis, Yinhan Liu, Naman Goyal, Marjan Ghazvininejad, Abdelrahman Mohamed, Omer Levy, Veselin Stoyanov, and Luke Zettlemoyer. 2020.
\newblock \href {https://doi.org/10.18653/v1/2020.acl-main.703} {{BART}: Denoising sequence-to-sequence pre-training for natural language generation, translation, and comprehension}.
\newblock In \emph{Proceedings of the 58th Annual Meeting of the Association for Computational Linguistics}, pages 7871--7880, Online. Association for Computational Linguistics.

\bibitem[{Lim et~al.(2017)Lim, Muis, Lu, and Ong}]{lim-etal-2017-malwaretextdb}
Swee~Kiat Lim, Aldrian~Obaja Muis, Wei Lu, and Chen~Hui Ong. 2017.
\newblock \href {https://doi.org/10.18653/v1/P17-1143} {{M}alware{T}ext{DB}: A database for annotated malware articles}.
\newblock In \emph{Proceedings of the 55th Annual Meeting of the Association for Computational Linguistics (Volume 1: Long Papers)}, pages 1557--1567, Vancouver, Canada. Association for Computational Linguistics.

\bibitem[{Liu et~al.(2019)Liu, Ott, Goyal, Du, Joshi, Chen, Levy, Lewis, Zettlemoyer, and Stoyanov}]{liu2019roberta}
Yinhan Liu, Myle Ott, Naman Goyal, Jingfei Du, Mandar Joshi, Danqi Chen, Omer Levy, Mike Lewis, Luke Zettlemoyer, and Veselin Stoyanov. 2019.
\newblock \href {http://arxiv.org/abs/1907.11692} {Roberta: {A} robustly optimized {BERT} pretraining approach}.
\newblock \emph{CoRR}, abs/1907.11692.

\bibitem[{Man Duc~Trong et~al.(2020)Man Duc~Trong, Trong~Le, Pouran Ben~Veyseh, Nguyen, and Nguyen}]{man-duc-trong-etal-2020-introducing}
Hieu Man Duc~Trong, Duc Trong~Le, Amir Pouran Ben~Veyseh, Thuat Nguyen, and Thien~Huu Nguyen. 2020.
\newblock \href {https://doi.org/10.18653/v1/2020.emnlp-main.433} {Introducing a new dataset for event detection in cybersecurity texts}.
\newblock In \emph{Proceedings of the 2020 Conference on Empirical Methods in Natural Language Processing (EMNLP)}, pages 5381--5390, Online. Association for Computational Linguistics.

\bibitem[{Mikolov et~al.(2013)Mikolov, Chen, Corrado, and Dean}]{mikolov2013word2vec}
Tom{\'{a}}s Mikolov, Kai Chen, Greg Corrado, and Jeffrey Dean. 2013.
\newblock \href {http://arxiv.org/abs/1301.3781} {Efficient estimation of word representations in vector space}.
\newblock In \emph{1st International Conference on Learning Representations, {ICLR} 2013, Scottsdale, Arizona, USA, May 2-4, 2013, Workshop Track Proceedings}.

\bibitem[{Pennington et~al.(2014)Pennington, Socher, and Manning}]{pennington-etal-2014-glove}
Jeffrey Pennington, Richard Socher, and Christopher Manning. 2014.
\newblock \href {https://doi.org/10.3115/v1/D14-1162} {{G}lo{V}e: Global vectors for word representation}.
\newblock In \emph{Proceedings of the 2014 Conference on Empirical Methods in Natural Language Processing ({EMNLP})}, pages 1532--1543, Doha, Qatar. Association for Computational Linguistics.

\bibitem[{Phandi et~al.(2018)Phandi, Silva, and Lu}]{phandi-etal-2018-semeval}
Peter Phandi, Amila Silva, and Wei Lu. 2018.
\newblock \href {https://doi.org/10.18653/v1/S18-1113} {{S}em{E}val-2018 task 8: Semantic extraction from {C}ybersec{U}rity {RE}ports using natural language processing ({S}ecure{NLP})}.
\newblock In \emph{Proceedings of The 12th International Workshop on Semantic Evaluation}, pages 697--706, New Orleans, Louisiana. Association for Computational Linguistics.

\bibitem[{Pujari et~al.(2021)Pujari, Friedrich, and Str{\"o}tgen}]{pujari2021multi}
Subhash~Chandra Pujari, Annemarie Friedrich, and Jannik Str{\"o}tgen. 2021.
\newblock A multi-task approach to neural multi-label hierarchical patent classification using transformers.
\newblock In \emph{European Conference on Information Retrieval}, pages 513--528. Springer.

\bibitem[{Rahman et~al.(2021)Rahman, Mahdavi{-}Hezaveh, and Williams}]{rahman2021what}
Md.~Rayhanur Rahman, Rezvan Mahdavi{-}Hezaveh, and Laurie~A. Williams. 2021.
\newblock \href {http://arxiv.org/abs/2109.06808} {What are the attackers doing now? automating cyber threat intelligence extraction from text on pace with the changing threat landscape: {A} survey}.
\newblock \emph{CoRR}, abs/2109.06808.

\bibitem[{Sanagavarapu et~al.(2021)Sanagavarapu, Iyer, and Reddy}]{sanagavarapu2021ontoenricher}
Lalit~Mohan Sanagavarapu, Vivek Iyer, and Y.~Raghu Reddy. 2021.
\newblock \href {http://arxiv.org/abs/2102.04081} {Ontoenricher: {A} deep learning approach for ontology enrichment from unstructured text}.
\newblock \emph{CoRR}, abs/2102.04081.

\bibitem[{Sarhan and Spruit(2021)}]{sarhan2021open}
Injy Sarhan and Marco Spruit. 2021.
\newblock \href {https://doi.org/https://doi.org/10.1016/j.knosys.2021.107524} {{Open-CyKG: An Open Cyber Threat Intelligence Knowledge Graph}}.
\newblock \emph{Knowledge-Based Systems}, 233:107524.

\bibitem[{Satyapanich et~al.(2020)Satyapanich, Ferraro, and Finin}]{Satyapanich_Ferraro_Finin_2020}
Taneeya Satyapanich, Francis Ferraro, and Tim Finin. 2020.
\newblock \href {https://doi.org/10.1609/aaai.v34i05.6401} {Casie: Extracting cybersecurity event information from text}.
\newblock \emph{Proceedings of the AAAI Conference on Artificial Intelligence}, 34(05):8749--8757.

\bibitem[{Saur{\i} et~al.(2006)Saur{\i}, Littman, Knippen, Gaizauskas, Setzer, and Pustejovsky}]{sauri2006timeml}
Roser Saur{\i}, Jessica Littman, Bob Knippen, Robert Gaizauskas, Andrea Setzer, and James Pustejovsky. 2006.
\newblock Timeml annotation guidelines version 1.2. 1.

\bibitem[{Simran et~al.(2020)Simran, Sriram, Vinayakumar, and Soman}]{simran2020deep}
K.~Simran, S.~Sriram, R.~Vinayakumar, and K.~P. Soman. 2020.
\newblock Deep learning approach for intelligent named entity recognition of cyber security.
\newblock In \emph{Advances in Signal Processing and Intelligent Recognition Systems}, pages 163--172, Singapore. Springer Singapore.

\bibitem[{Str{\"o}tgen and Gertz(2010)}]{strotgen-gertz-2010-heideltime}
Jannik Str{\"o}tgen and Michael Gertz. 2010.
\newblock \href {https://aclanthology.org/S10-1071} {{H}eidel{T}ime: High quality rule-based extraction and normalization of temporal expressions}.
\newblock In \emph{Proceedings of the 5th International Workshop on Semantic Evaluation}, pages 321--324, Uppsala, Sweden. Association for Computational Linguistics.

\bibitem[{UzZaman et~al.(2013)UzZaman, Llorens, Derczynski, Allen, Verhagen, and Pustejovsky}]{uzzaman-etal-2013-semeval}
Naushad UzZaman, Hector Llorens, Leon Derczynski, James Allen, Marc Verhagen, and James Pustejovsky. 2013.
\newblock \href {https://aclanthology.org/S13-2001} {{S}em{E}val-2013 task 1: {T}emp{E}val-3: Evaluating time expressions, events, and temporal relations}.
\newblock In \emph{Second Joint Conference on Lexical and Computational Semantics (*{SEM}), Volume 2: Proceedings of the Seventh International Workshop on Semantic Evaluation ({S}em{E}val 2013)}, pages 1--9, Atlanta, Georgia, USA. Association for Computational Linguistics.

\bibitem[{Wu et~al.(2020)Wu, Petroni, Josifoski, Riedel, and Zettlemoyer}]{wu-etal-2020-scalable}
Ledell Wu, Fabio Petroni, Martin Josifoski, Sebastian Riedel, and Luke Zettlemoyer. 2020.
\newblock \href {https://doi.org/10.18653/v1/2020.emnlp-main.519} {Scalable zero-shot entity linking with dense entity retrieval}.
\newblock In \emph{Proceedings of the 2020 Conference on Empirical Methods in Natural Language Processing (EMNLP)}, pages 6397--6407, Online. Association for Computational Linguistics.

\bibitem[{Zhu et~al.(2015)Zhu, Kiros, Zemel, Salakhutdinov, Urtasun, Torralba, and Fidler}]{zhu2015googlebooks}
Yukun Zhu, Ryan Kiros, Richard~S. Zemel, Ruslan Salakhutdinov, Raquel Urtasun, Antonio Torralba, and Sanja Fidler. 2015.
\newblock \href {https://doi.org/10.1109/ICCV.2015.11} {Aligning books and movies: Towards story-like visual explanations by watching movies and reading books}.
\newblock In \emph{2015 {IEEE} International Conference on Computer Vision, {ICCV} 2015, Santiago, Chile, December 7-13, 2015}, pages 19--27. {IEEE} Computer Society.

\end{thebibliography}

\clearpage
\appendix
\begin{center}
\bfseries \Large{Appendix}
\end{center}

\section{Preprocessing}
\label{sec:appendix-vendors}

The texts were retrieved using Python Requests\footnote{\href{https://github.com/psf/requests}{https://github.com/psf/requests}}.
Due to the URLs, code and image references within the texts, we convert the articles to a format similar to Markdown using BeautifulSoup\footnote{\href{https://www.crummy.com/software/BeautifulSoup/}{https://www.crummy.com/software/BeautifulSoup/}} and Markdownify\footnote{\href{https://github.com/matthewwithanm/python-markdownify}{github.com/matthewwithanm/python-markdownify}}.
Because off-the-shelf sentence segmenters do not perform well on texts that contain many links or code snippets, we use a custom regular expression tokenizer for sentence segmentation and correct sentence boundaries manually before uploading the texts to the web-based annotation system INCEpTION \citep{klie-etal-2018-inception}.

\section{Corpus Statistics}
\label{sec:appendix-corpus-stats}

\begin{figure*}[!hbp]
    \centering
    \includegraphics[width=\textwidth]{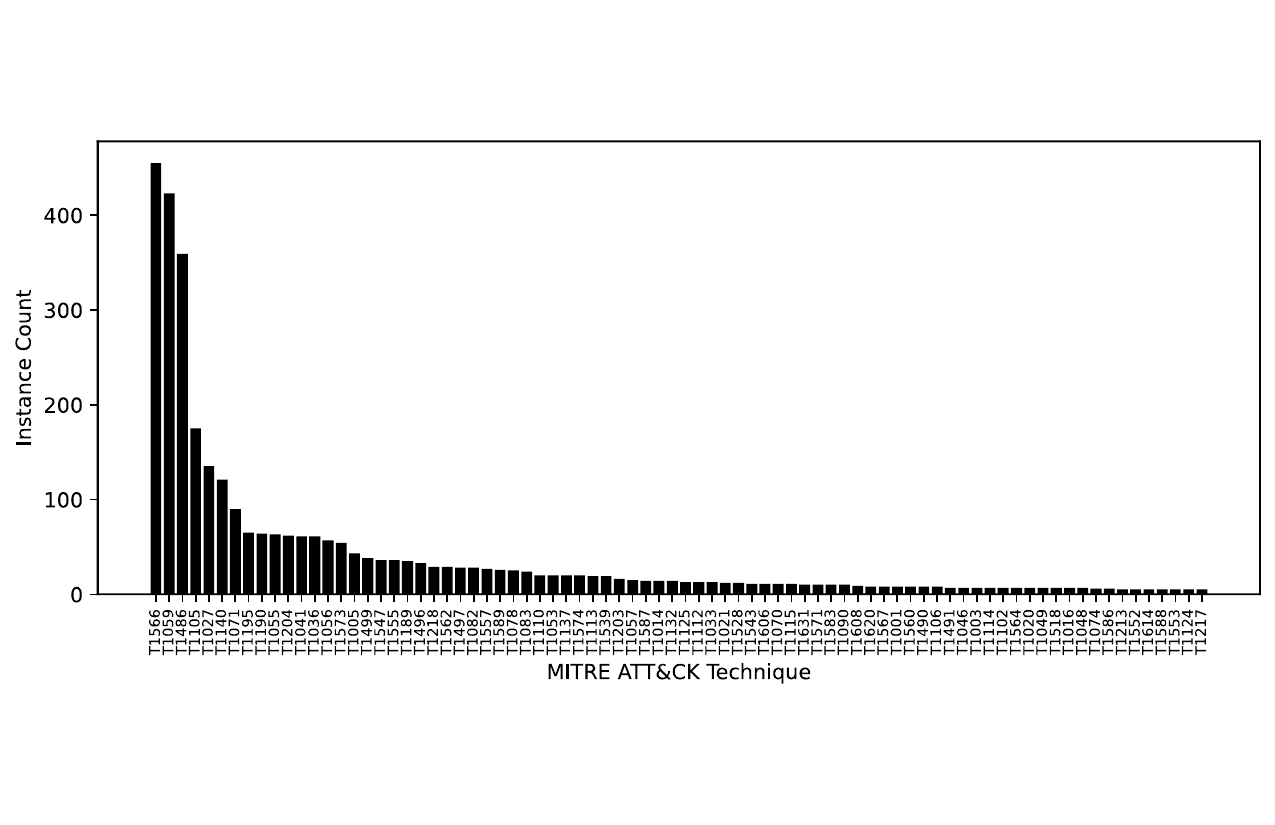}
    \caption{Distribution of technique links annotated in \corpusName, showing techniques that occur at least 5 times. In addition, there is a long tail of 136 instances annotated with techniques occurring less frequently.}
    \label{fig:technique-dist}
\end{figure*}

\begin{figure*}[!hbp]
    \centering
    \includegraphics[width=0.7\textwidth]{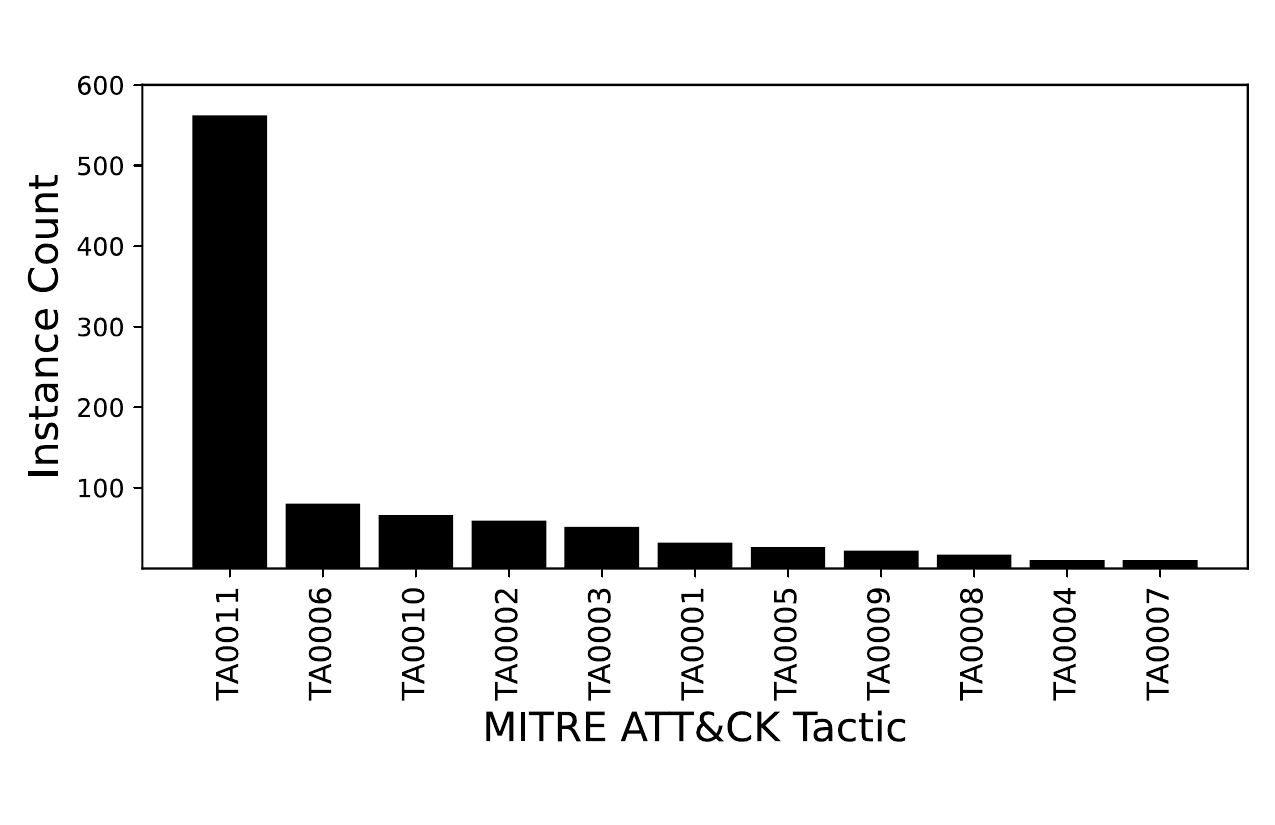}
    \caption{Distribution of tactic links annotated in \corpusName.}
    \label{fig:tactic-dist}
\end{figure*}

\fref{fig:technique-dist} and \fref{fig:tactic-dist} show the distributions over technique and tactic annotations in \corpusName, respectively.

\tref{tab:corpus_splits_app} shows details on the training/dev/test splits of our corpus grouped by CTI vendor.

\section{NER results per class}
\label{sec:appendix-ner-details}
The detailed NER results per class are shown in \tref{tab:ner-results-per-class}. 

\section{Computational experiments}
\label{sec:appendix-hyperparameters}
In the following, we report further information on our computational experiments. 

\paragraph{Computing infrastructure.}
We use V100 gpus for all experiments with neural models. 

\paragraph{Number of parameters.}
The base-sized transformer models (BERT, SciBERT, CodeBERT, RoBERTa) have 110M parameters and take 20 minutes to train for NER. 
The two BERT-large model used in BLINK the BLINK encoder have 340M parameters each and take 10 hours to train. 
The GENRE model has 406M parameters and train for 12 hours. 
The BiLSTM-CRF baseline for NER has 41M parameters and takes 2 hours to train.  
The TMM model has 145M parameters and takes 1 hour to train.

\begin{table*}
\footnotesize
\centering
\begin{tabular}{ll|c|c|c|c|c|c} \toprule
                            &            & Intel471   & Lab52      & ProofPoint & QuoIntelligence & ZScalar    & total \\ \midrule
\multirow{3}{*}{\begin{tabular}[c]{@{}l@{}}Train\\ (GNE+CyNE)\end{tabular}}      & \#docs.    & 18         & 13         & 16         & 7                 & 16         & 70    \\
                            & Start date & 2016-06-17 & 2019-04-02 & 2021-04-15 & 2018-11-29        & 2020-06-26 &       \\
                            & End date   & 2021-04-07 & 2002-04-14 & 2021-08-31 & 2021-01-16        & 2021-05-11 &       \\ \midrule
\multirow{3}{*}{\begin{tabular}[c]{@{}l@{}}Train add.$^{X}$\\ (GNE)\end{tabular}}& \#docs.    & 4          & 5          & 73         & 2                 & 196        & 280   \\
                            & Start date & 2020-04-01 & 2019-04-02 & 2017-11-29 & 2020-01-27        & 2013-03-08 &       \\
                            & End date   & 2021-10-20 & 2021-12-14 & 2022-02-15 & 2020-07-20        & 2021-07-20 &       \\ \midrule
\multirow{3}{*}{\begin{tabular}[c]{@{}l@{}}Dev\\ (GNE+CyNE)\end{tabular}}        & \#docs.    & 4          & 3          & 4          & 1                 & 4          & 16    \\
                            & Start date & 2021-04-19 & 2020-05-14 & 2021-09-29 & 2021-01-27        & 2021-05-21 &       \\
                            & End date   & 2021-05-15 & 2020-06-09 & 2021-10-27 & 2021-01-27        & 2021-07-14 &       \\ \midrule
\multirow{3}{*}{\begin{tabular}[c]{@{}l@{}}Test\\ (GNE+CyNE)\end{tabular}}       & \#docs.    & 8          & 7          & 8          & 4                 & 7          & 34    \\
                            & Start date & 2021-07-14 & 2020-08-26 & 2021-10-28 & 2021-02-16        & 2021-09-09 &       \\
                            & End date   & 2021-12-09 & 2022-01-24 & 2022-02-03 & 2021-06-30        & 2021-11-16 &       \\ \bottomrule
\end{tabular}
\caption{Detailed information on datasplits.}
\label{tab:corpus_splits_app}
\end{table*}

\newcommand{\ba}[1]{\textbf{#1}}
\newcommand{\bb}[1]{\textit{#1}}

\begin{table*}[b]
\setlength\tabcolsep{3pt}
\footnotesize
\centering
\begin{tabular}{lcc|cc|cc|cccc} \toprule
            & BERT              & $^{+X}$           & SciBERT      & $^{+X}$           & CodeBERT     & $^{+X}$           & \multicolumn{3}{c}{RoBERTa}                          & $^{+X}$      \\ 
            & F1                & F1                & F1           & F1                & F1           & F1                & Pre.         & Rec.         & F1                     & F1           \\ \midrule
CodeSnippet & 37.6\sd{3.9}      & 41.7\sd{5.0}      & 34.5\sd{5.6} & \bb{42.0\sd{6.3}} & 28.9\sd{6.7} & 37.1\sd{5.3}      & 30.7\sd{4.2} & 53.3\sd{5.6} & \ba{38.9\sd{4.7}}      & 37.0\sd{3.6} \\
DATE        & \ba{86.8\sd{1.1}} & 86.6\sd{1.0}      & 85.9\sd{1.1} & 86.2\sd{0.9}      & 85.3\sd{0.8} & \bb{87.2\sd{1.2}} & 81.8\sd{1.7} & 87.2\sd{1.2} & 84.4\sd{1.2}           & 86.8\sd{0.7} \\
LOC         & 82.5\sd{1.0}      & \bb{86.6\sd{1.4}} & 80.4\sd{0.5} & 84.5\sd{0.6}      & 81.2\sd{1.9} & 84.2\sd{2.5}      & 83.2\sd{2.4} & 84.7\sd{0.5} & \ba{83.9\sd{1.1}}      & 86.2\sd{0.9} \\
ORG         & 80.9\sd{2.0}      & 86.4\sd{0.8}      & 79.1\sd{2.1} & 85.6\sd{0.8}      & 80.4\sd{1.9} & 86.4\sd{0.8}      & 81.2\sd{1.5} & 91.0\sd{1.1} & \ba{85.8\sd{1.2}}      & \bb{88.5\sd{1.1}} \\
SECTOR      & 52.8\sd{2.3}      & 55.1\sd{2.1}      & 56.3\sd{3.2} & \bb{60.4\sd{3.5}} & 50.7\sd{3.2} & 55.2\sd{4.5}      & 56.4\sd{3.5} & 56.1\sd{3.8} & \bb{\ba{60.4\sd{3.4}}} & 63.3\sd{2.1} \\ \midrule
CON         & 61.8\sd{1.0}      & -                 & 63.4\sd{2.6} & -                 & 64.6\sd{1.3} & -                 & 69.5\sd{2.8} & 66.8\sd{1.4} & \ba{68.1\sd{1.2}}      & -         \\
GROUP       & 43.6\sd{3.6}      & -                 & 45.6\sd{2.6} & -                 & 40.5\sd{3.5} & -                 & 70.6\sd{5.7} & 37.3\sd{1.0} & \ba{48.7\sd{1.6}}      & -         \\
MALWARE     & 58.9\sd{3.6}      & -                 & 62.9\sd{2.9} & -                 & 65.5\sd{5.0} & -                 & 69.1\sd{2.7} & 69.6\sd{5.1} & \ba{69.3\sd{3.2}}      & -         \\
TOOL        &  8.1\sd{2.3}      & -                 & 16.6\sd{3.3} & -                 & \ba{18.0\sd{15.2}} & -           &  8.5\sd{4.5} &  8.1\sd{2.8} & 8.2\sd{3.5}            & -         \\
TACTIC      & 54.1\sd{2.9}      & -                 & 57.0\sd{2.0} & -                 & 54.7\sd{3.4} & -                 & 60.0\sd{4.3} & 57.5\sd{3.8} & \ba{58.6\sd{3.4}}      & -         \\
TECHNIQUE   & 37.1\sd{1.3}      & -                 & 36.0\sd{0.8} & -                 & 38.5\sd{2.8} & -                 & 39.8\sd{5.0} & 42.9\sd{1.6} & \ba{41.2\sd{3.3}}      & -         \\ \bottomrule
\end{tabular}
\caption{Sequence Labeling results (average Micro F1 and the standard deviation of 5 runs) We report recall and precision for the best model (RoBERTa). $^{+X}$ marks models trained with extra data. }
\label{tab:ner-results-per-class}
\end{table*}

\end{document}